\pdfoutput=1

\documentclass[11pt]{article}

\usepackage[final]{acl}

\usepackage{times}
\usepackage{latexsym}

\usepackage[T1]{fontenc}

\usepackage[utf8]{inputenc}

\usepackage{microtype}

\usepackage{inconsolata}

\newcommand{\eat}[1]{}

\usepackage[textwidth=0.7in]{todonotes}

\newcommand{\modelname}{Design2Code\xspace}

\usepackage[utf8]{inputenc} 
\usepackage[T1]{fontenc}    
\usepackage{hyperref}       
\usepackage{url}            
\usepackage{booktabs}       
\usepackage{amsfonts}       
\usepackage{nicefrac}       
\usepackage{microtype}      
\usepackage{xcolor}         

\usepackage[utf8]{inputenc} 
\usepackage[T1]{fontenc}    
\usepackage{hyperref}       
\usepackage{url}            
\usepackage{booktabs}       
\usepackage{amsfonts}       
\usepackage{xspace}
\usepackage{nicefrac}       
\usepackage{microtype}      
\usepackage{xcolor}         
\usepackage{natbib}
\usepackage{amsmath}
\usepackage{comment}
\usepackage{siunitx}
\usepackage{graphicx}
\usepackage{adjustbox}
\usepackage{wrapfig}
\usepackage{caption}
\usepackage[most]{tcolorbox}
\usepackage{wrapfig}
\newtcolorbox{myquote}{
    breakable,
    width=0.48\textwidth,
    colback=white,
    colframe=black,
    fontupper=\itshape,
    boxrule=0.2mm,
    left=1mm,
    right=1mm,
    arc=3mm,
    auto outer arc
}

\title{\texttt{\modelname:}  Benchmarking Multimodal Code Generation for Automated Front-End Engineering}

\author{Chenglei Si\thanks{\ \ Equal contribution.}$^1$, Yanzhe Zhang$^{*2}$, Ryan Li$^1$, Zhengyuan Yang$^3$, Ruibo Liu$^4$, Diyi Yang$^1$ \\
   $^1$Stanford University, $^2$Georgia Tech, $^3$Microsoft, $^4$Google DeepMind\\
  \texttt{clsi@stanford.edu, z\_yanzhe@gatech.edu}
  }

\begin{document}
\maketitle
\begin{abstract}
Generative AI has made rapid advancements in recent years, achieving unprecedented capabilities in multimodal understanding and code generation. This can enable a new paradigm of front-end development in which multimodal large language models (MLLMs) directly convert visual designs into code implementations. In this work, we construct
\texttt{\modelname} -- the first real-world benchmark for this task. Specifically, we manually curate 484 diverse real-world webpages as test cases and develop a set of automatic evaluation metrics to assess how well current multimodal LLMs can generate the code implementations that directly render into the given reference webpages, given the screenshots as input. We also complement automatic metrics with comprehensive human evaluations to validate the performance ranking. To rigorously benchmark MLLMs, we test various multimodal prompting methods on frontier models such as GPT-4o, GPT-4V, Gemini, and Claude. Our fine-grained break-down metrics indicate that models mostly lag in recalling visual elements from the input webpages and generating correct layout designs.
\end{abstract}

\section{Introduction}

\begin{figure*}[t]
  \centering
  \includegraphics[trim={1cm 1cm 4cm 1cm},width=2.0\columnwidth]{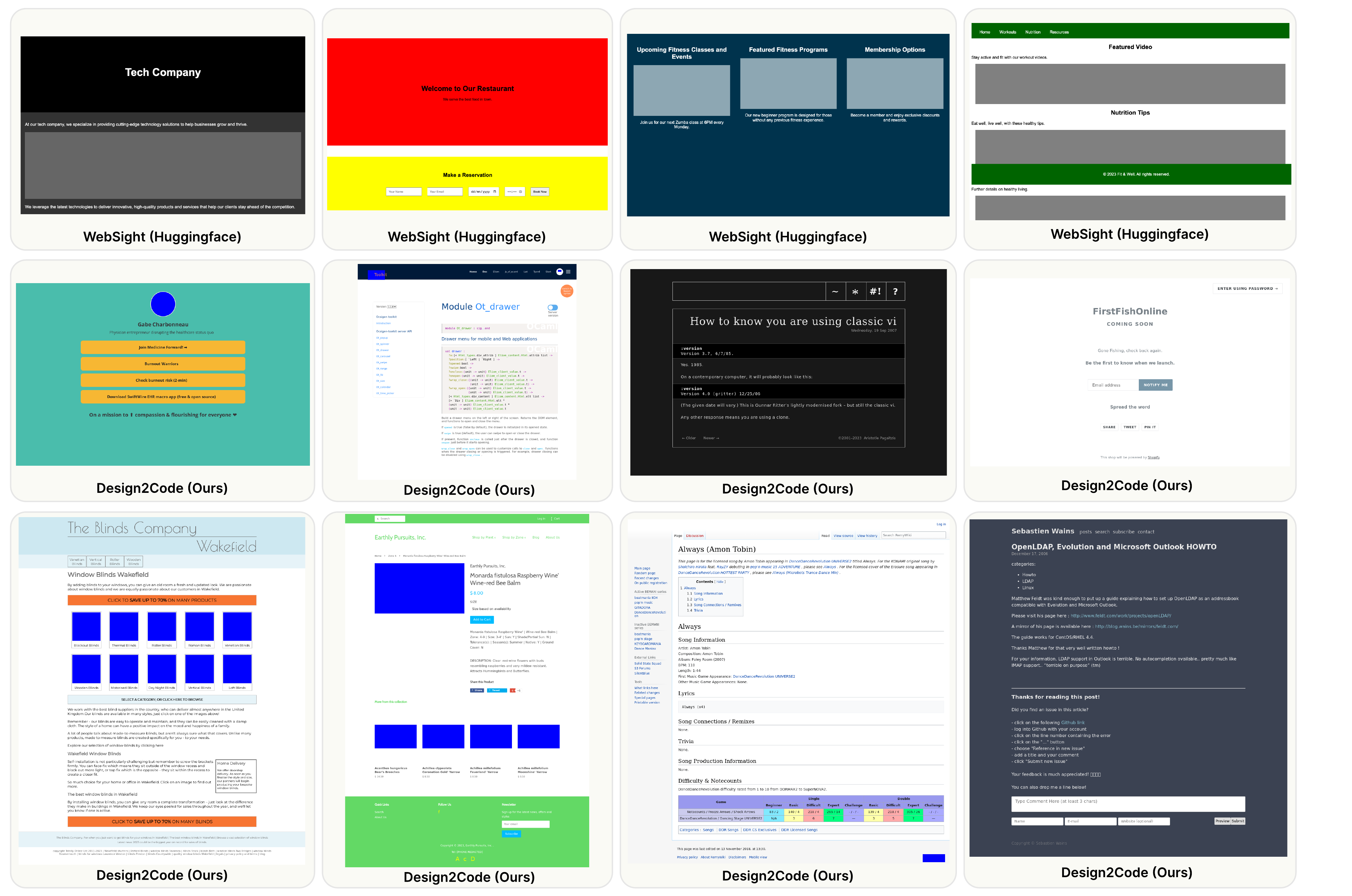}
  \caption{Examples from the prior WebSight v0.1 dataset (first row) and our new \modelname benchmark (last two rows).
  We use real-world webpages for benchmarking to ensure they are realistic, diverse, and complex, while WebSight uses synthetically generated simple webpages for scalability.
  }
  \label{fig:test_examples}
\end{figure*}


Implementing visual designs of websites into functional code is a challenging task as it requires understanding visual elements and their layouts and then translating them into structured code.
Such dependencies on sophisticated skills have prevented many laypeople from building their own web applications, even when they have concrete ideas for what to build. 
Furthermore, the requirement for domain expertise complicates the whole webpage production pipeline, requiring collaboration among people with different skill sets and potentially causing discrepancies between the intended design and actual implementation. 
Effective automatic generation of functional code from visual designs 
 has the potential to democratize the development of front-end web applications~\citep{nguyen2015reverse}, allowing non-experts to build applications easily and quickly.
%

While code generation from natural language instructions has advanced rapidly in recent years \citep{yin2017syntactic,le2020deep,StarCoder}, generating code implementation from user interface (UI) design has not received much attention due to a wide range of challenges, such as diversity in visual and text signals on the user interface and the vast search space in the resulting code.  
\citet{beltramelli2018pix2code} made a notable attempt back in 2017 with CNN and RNN models on a narrow set of simplistic UI designs. Over the years, despite many follow-up attempts along this quest~\citep{Robinson2019Sketch2codeGA,Soselia2023LearningUR}, they are all constrained to simplistic or synthetic examples with a narrow set of layout designs, hardly useful for real-world front-end development applications. Until recently, the development of multimodal LLMs has entered a new era where large-scale pretrained models can process both visual and text input and generate text output for various visually grounded tasks, with representative examples like Flamingo~\citep{Alayrac2022FlamingoAV}, GPT-4V~\citep{GPT4V}, and Gemini~\citep{Gemini}. Such advancement has unlocked a brand new paradigm for this long-standing unsolved task:
 give the user's website design as an image input to the system to obtain the full code implementation that can render into the desired webpage in an end-to-end manner. 
 To systematically and rigorously benchmark current MLLMs on this task, we construct the first-ever real-world benchmark \textbf{\texttt{\modelname}}.


To best reflect realistic use cases, we use real-world webpages (examples in Figure~\ref{fig:test_examples}) in the wild as our test examples rather than synthetically generated ones as in prior works~\citep{Soselia2023LearningUR, laurencon2024unlocking}. 
%
%
We scrape webpages in the C4~\citep{2019t5} validation set and perform careful manual curation to obtain a set of 484 high-quality, challenging, and diverse webpages representing a wide variety of real-world use cases with different levels of complexities. 
We show both quantitatively and qualitatively that our benchmark covers a wide spectrum of HTML tag uses, domains, and complexity levels. 
To facilitate efficient evaluation and model development, we also develop automatic metrics for this task that measure the \emph{visual design similarity} between the generated webpage's screenshot and the given screenshot input. Our metrics consider a comprehensive set of dimensions, including bounding box matches, text content, position, and color of all matched visual elements on the webpages, which we later show highly correlate with human judgment.

We then investigate how current MLLMs perform on this task, 
by testing a variety of prompting methods, including our text-augmented prompting that complements visual input with extracted text elements from the webpage to reduce the load on OCR, as well as a self-revision prompting method that asks the model to compare its previous generation and the input webpage for self-improvement. 
We see significant improvement from text-augmented prompting on most models, while few can self-revise their generation.
Additionally, we also construct \textbf{\texttt{\modelname-HARD}}, a separate set of 80 hard examples, to compare state-of-the-art commercial models (GPT-4o, Claude 3.5 Sonnet) on these challenging cases.

\section{The \texttt{\modelname} Benchmark}
\label{sec: design2code benchmark}

In this section, we describe the curation and processing of our benchmark data. We first scrape all website links in the C4~\citep{2019t5} validation set. We then embed all CSS code into the HTML file to obtain one single code implementation file for each webpage. This results in a total of 127.9k webpages, which we perform further filtering and processing as described below.

\begin{figure}[t]
  \centering
\includegraphics[width=0.96\columnwidth]{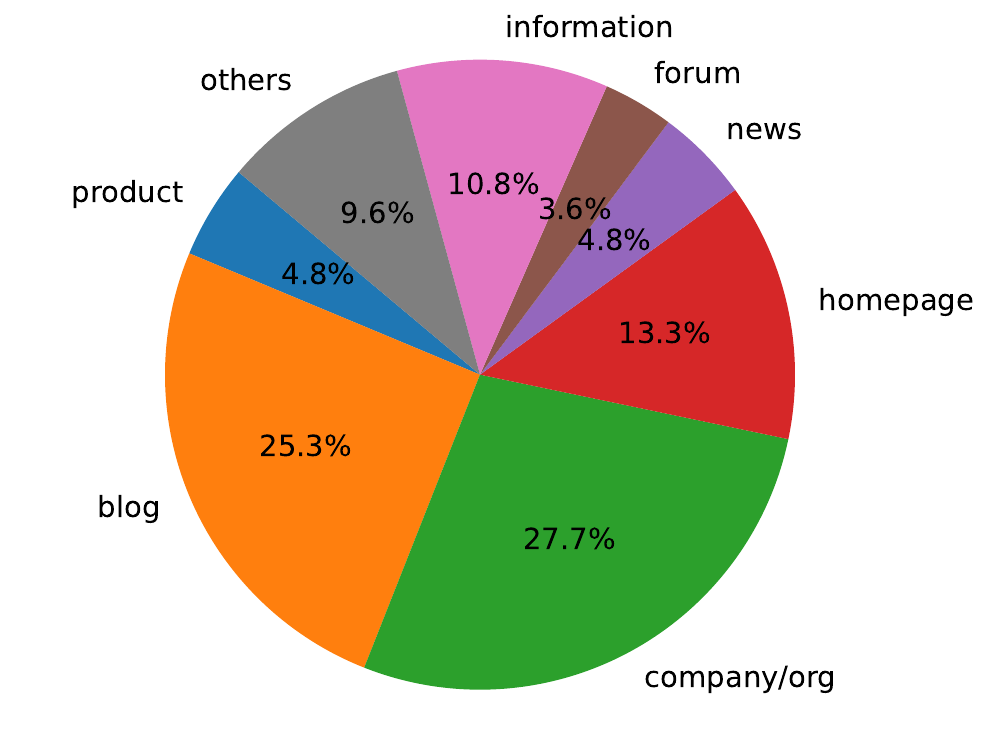}
  \caption{Main topics of the webpages in the \modelname benchmark.}
  \label{fig:topic_pie_chart}
\end{figure}

\subsection{Test Set Curation}

Our overall goal is to obtain a set of well-formed webpages that represent diverse real-world use cases. We follow the following steps for automatic processing and manual filtering: First, we automatically filter out webpages that are too long or too simple (only contain images or texts) and run deduplication, which results in 14k webpages remaining. We then remove external dependencies and replace multimedia content with placeholders. Finally, the first two authors conduct manual curation to check the independence from external files, the absence of sensitive content, and proper formatting. This quality filtering narrows down the samples to 484 high-quality webpages, which are used for our benchmark. More data processing details are described in Appendix Sec \ref{sec: detailed test set curation}.

\subsection{Data Statistics and Diversity}

\noindent
\textbf{Quantitative Metrics} To provide an estimate of the difficulty levels of the test examples, we provide some quantitative measures. 
\textbf{(1) Length}: We tokenize the scraped code files with the \texttt{GPT-2} tokenizer. The average number of tokens per file is 31216 (min=784, max=98637, std=23903). This is much longer than the typical max output length of modern language models, posing a unique challenge.
\textbf{(2) Total number of tags}: We count the total number of HTML tags involved, which is 158 on average (min=12, max=528, std=100). 
The examples in our benchmark cover 84 types of standard HTML5 tags. 
We present a chart of the most frequently used tags in Appendix Table~\ref{tab:frequency_tags}. \textbf{(3) DOM tree depth}: We measure the depth of the Document Object Model (DOM) tree as another measure of complexity. The average depth is 13 (min=4, max=32, std=5). 
\textbf{(4) Number of unique tags}: Lastly, we compute the number of unique HTML tags in each example, and the mean is 22 (min=8, max=45, std=6), suggesting that our benchmark covers a wide range of HTML tags. 
We compare these metrics with the most recent and most similar existing dataset -- WebSight~\citep{laurencon2024unlocking} in Appendix Table~\ref{tab:dataset_comparison}.
Overall, the examples in our benchmark are much more challenging and cover a wider spectrum of complexities than prior efforts like WebSight.

\noindent
\textbf{Domain Distribution} To get a sense of the range of domains covered in our benchmark, we randomly sample 25\% examples (N=120) from the benchmark and manually annotate what type of webpages they are based on their functions. We present the pie chart of the most frequent domains in Figure~\ref{fig:topic_pie_chart}. The most prominent genres are websites of companies or organizations, personal blogs (including technical blogs), and personal homepages. Other genres include information-sharing sites (e.g., Wikipedia pages, FAQ pages, tax policy pages, online dictionaries), online forums, news article pages, and product description pages. Sampled examples are shown in Figure~\ref{fig:test_examples}.


\subsection{Automatic Metrics}

Previously, auto-generated HTML code is usually evaluated by text-based similarity metrics, such as Normalized Edit Distance \citep{kosmos25} and htmlBLEU \citep{Soselia2023LearningUR}. However, such metrics cannot directly assess whether the visual design of the original screenshot is correctly generated as there can be many different ways of implementing the same webpage, and minor differences in generated code could result in major visual differences in the rendered output. 
To this end, we propose to automatically evaluate generated webpages by calculating the similarity between the screenshots of reference webpages $I_R$ and the rendered screenshots of generated webpages $I_G$. 
We break down the evaluation into both high-level visual similarity and low-level element matching. 

\noindent
\textbf{High-level Visual Similarity} To evaluate the visual similarity of $I_R$ and $I_G$, we use the similarity of their CLIP \citep{CLIP} embedding, denoted as $\mathbf{CLIP}(I_R, I_G)$. Specifically, we extract features by \texttt{CLIP-ViT-B/32} after resizing screenshots to squares. To rule out the texts in the screenshots, we use the inpainting algorithm from \citet{inpainting} to mask all detected text boxes using their bounding box coordinates.~\footnote{\url{https://docs.opencv.org/4.3.0/df/d3d/tutorial_py_inpainting.html}}

\noindent
\textbf{Low-level Element Matching}
Metrics like CLIP similarity only capture the similarity of the overall images rather than the matching of all the details like text. Moreover, the metric itself does not offer any fine-grained breakdown to help diagnose model weaknesses. To complement that, we introduce a suite of element-matching metrics. 
Specifically, we consider whether the generated webpages manage to recall all visual elements, 
and whether the corresponding visual elements in the reference and generated webpages have aligned text content,  position, and color \citep{cao20155, still2018web}. 

Given a reference webpage screenshot $I_R$ and a generated webpage screenshot $I_G$, we use a text detection module to output a set of detected visual element blocks for each: $R = \{r_1, r_2,..., r_m\}$ and $G = \{g_1, g_2,..., g_n\}$, where each block contains its textual content and bounding box coordinates. See Appendix~\ref{sec:detection} for the details of implementing the block detection module.
Based on the two sets of detected blocks, we use the Jonker-Volgenant algorithm \citep{matchingalgo} to get the optimal matching $M$ between $R$ and $G$ based on text similarity, where $(p, q) \in M$ indicates $r_p$ is matched with $g_q$. Given $R$, $G$, and matched pairs in $M$, we evaluate similarity along the following aspects:

\noindent
\textbf{$*$ Block-Match}: 
The first desideratum of the task is that all visual elements from the reference webpage should be reproduced in the generated webpage, and the generated webpage should not hallucinate non-existent new elements. 
We measure this by computing the total sizes of all matched blocks divided by the total sizes of all blocks (Equation (1)(2) in Figure \ref{equations}), including unmatched ones (either because the generated webpages missed them or because the generated webpages contain hallucinated blocks).
\begin{figure*}[t]
\centering
\begin{align*}
    \mathbf{match_{block}}(r_p, g_q) &= \frac{ S(r_p) + S(g_q)}{\sum_{(i,j) \in M}  (S(r_i) + S(g_j)) + (\sum_{i \in U_R} S(r_i) + \sum_{j \in U_G} S(g_j))} \tag{1}, \\
    \mathbf{match_{block}}(R, G) &= \sum_{(p,q) \in M} \mathbf{match_{block}}(r_p, g_q) \tag{2}.
\end{align*}
\caption{\label{equations}Equations to calculate the Block-Match metric.}
\end{figure*}
$S(\cdot)$ returns the size of the blocks, $U_R$ and $U_G$ denotes the unmatched blocks in $R$ and $G$.
The intuition is that unmatched blocks will lower the score as they indicate missing original blocks or generating hallucinated blocks; the larger the unmatched blocks are, the lower this score is. 

\noindent
\textbf{$*$ Text}: 
Given two strings from two matched blocks $r_p$ and $g_q$, the text similarity $\mathbf{sim_{text}}(r_p, g_q)$ is calculated as twice the number of overlapping characters divided by the total number of characters in the two strings (character-level Sørensen-Dice similarity). The overall score is averaged across all matched pairs. 

\noindent
\textbf{$*$ Position}: 
The position of the blocks largely impacts the overall layout. 
For each matched pair $(p, q)$, we calculate the position similarity $\mathbf{sim_{pos}}(r_p, g_q) = 1 - max(abs(x_q - x_p), abs(y_q - y_p))$, where $(x_p, y_p)$ and $(x_q, y_q)$ are normalized coordinates (in $[0, 1]$) of $r_p$ and $g_q$'s centers. The overall score is averaged across all matched pairs. 

\noindent
\textbf{$*$ Color}: 
We use the CIEDE2000 color difference formula \citep{CIEDE2000} to assess the perceptual difference between the colors of the generated text in block $g_q$ and the reference text in block $r_p$, denoted as $\mathbf{sim_{color}}(r_p, g_q))$, where the formula considers the complexities of human color vision. The overall score is averaged across all matched pairs. 

These low-level matching scores are designed as fine-grained diagnostic scores to complement the CLIP score. Ideally, models and methods should score well along all these dimensions.

\section{Benchmarking: Prompting and Finetuning}
\label{sec: prompting and finetuning}

We benchmark a variety of models and methods to compare their performance on our benchmark, including commercial API models, open-source models, and finetuned models.

\subsection{Prompting Methods}

We test a suite of multimodal prompting methods for our benchmark. We assume access to a model that can take both image input and text prompts and produce code as output. 

\noindent
\textbf{Direct Prompting}
We provide the reference webpage screenshot, along with the instruction to generate the HTML and CSS code (full prompt in Appendix~\ref{sec:prompts}).

\noindent
\textbf{Text-Augmented Prompting}
Direct prompting asks the model to do everything at once: recognize all the text and layout elements and generate the corresponding code. In reality, users often know what content they want to put on their webpage. Instead, they only seek expertise in converting the design into code implementation. To reflect such a setting, we also explore a text-augmented prompting method, where we extract all text elements from the original webpage first
and append these texts after the instruction prompt along with the screenshot input. In this setting, we mitigate the difficulty of performing OCR and instead allow the model to focus more on layout design, where the model could copy text content from the prompt and insert it into the correct positions.

\noindent
\textbf{Self-Revision Prompting}
To test whether the models can visually contrast the websites rendered by their generated code and the reference websites to further improve the code generation, we develop a self-revision prompt where we provide the following as input: (1) the screenshot of the input webpage,
(2) the screenshot of the generated webpage from text-augmented prompting,
(3) the generated code from text-augmented prompting as the initial solution;
then we ask the model to improve the generated implementation code so that the result can look closer to the reference webpage (full prompt is in Appendix~\ref{sec:prompts}).

\subsection{Model Setup}
We test the three prompting methods on the following commercial models: GPT-4o
, GPT-4V
~\citep{GPT4V}, Claude 3 Opus \citep{claude3}, Gemini 1.0 Pro Vision 
, 
as well as the following open models: LLaVA-V1.6-Mixtral-7B \citep{liu2024llavanext}, DeepSeek-VL-7B \citep{lu2024deepseekvl}, Idefics2-8B~\citep{Laurencon2024WhatMW}.
For finetuned models, we test WebSight VLM-8B, which is finetuned from Idefics2-8B on the full WebSight v0.1 dataset \citep{laurencon2024unlocking}, and our own finetuned \texttt{\modelname}-18B, a finetuned CogAgent-18B~\citep{cogagent} on a random subset (20\%) of the WebSight (technical finetuning details are in Appendix Sec \ref{sec: finetuning details}). 


\section{Results and Analysis}
\label{sec: Results and Analysis}

\subsection{Automatic Evaluation}

We present all automatic evaluation results in Table~\ref{tab:auto_eval}. 
Note that the comparisons here are not apple-to-apple comparisons, given the differences in model sizes and training data. We compare them as they are
\begin{table}[t]
\centering
\captionsetup{font=footnotesize}
\footnotesize
\centering
\begingroup
\setlength{\tabcolsep}{2pt} 
\renewcommand{\arraystretch}{1.0} 
\begin{tabular}{l|cccc|c}
\toprule
& Block & Text & Position & Color & CLIP  \\ 
\midrule
\multicolumn{6}{c}{GPT-4o} \\
\midrule 
Direct & \textbf{93.0} & 98.2 & \textbf{85.5} & \textbf{84.1} & \textbf{90.4}\\ 
Text-Augmented & 92.4 & \textbf{98.6} & 84.5 & 83.1 & 89.9 \\ 
Self-Revision & 92.7 & \textbf{98.6} & 84.9 & 83.3 & 90.1 \\
\midrule
\multicolumn{6}{c}{GPT-4V} \\
\midrule 
Direct & 85.8 & 97.4 & 80.5 & 73.3 & 86.9\\ 
Text-Augmented & 87.6 & 98.2 & 80.2 & 73.0 & 87.2\\ 
Self-Revision & 88.8 & 98.1 & 81.1 & 72.9 & 87.2  \\
\midrule
\multicolumn{6}{c}{Claude 3 Opus} \\
\midrule 
Direct & 90.2 & 97.5 & 77.9 & 71.4 & 87.0\\ 
Text-Augmented & 89.8 & 98.0 & 75.9 & 69.3 & 86.6\\ 
Self-Revision & 90.3 & 98.1 & 78.1 & 69.7 & 86.6  \\
\midrule
\multicolumn{6}{c}{Gemini 1.0 Pro Vision} \\
\midrule 
Direct & 80.2 & 94.6 & 72.3 & 66.2 & 84.4 \\ 
Text-Augmented & 84.8 & 96.9 & 70.4 & 66.3 & 84.4 \\ 
Self-Revision & 84.1 & 96.6 & 70.1 & 66.2 & 84.3 \\
\midrule 
\multicolumn{6}{c}{LLaVA 1.6-7B} \\
\midrule 
Direct & 50.4 & 87.9 & 69.1 & 63.4 & 84.6 \\ 
Text-Augmented & 68.4 & 93.0 & 68.7 & 64.0 & 84.5 \\ 
Self-Revision & 62.6 & 91.0 & 64.7 & 62.6 & 83.8 \\
\midrule 
\multicolumn{6}{c}{DeepSeek-VL-7B} \\
\midrule 
Direct & 39.7 & 77.0 & 64.6 & 63.8 & 84.5 \\ 
Text-Augmented & 66.1 & 93.4 & 69.2 & 67.9 & 84.3 \\ 
Self-Revision & 30.1 & 38.9 & 28.9 & 28.1 & 79.9 \\
\midrule 
\multicolumn{6}{c}{Idefics2-8B} \\
\midrule 
Direct & 46.7 & 80.3 & 55.9 & 58.9 & 81.7 \\ 
Text-Augmented & 23.6 & 55.6 & 35.7 & 36.3 & 78.7 \\ 
Self-Revision & 12.3 & 22.6 & 13.2 & 14.5 & 78.4 \\
\midrule 
\multicolumn{6}{c}{Finetuned Models}\\
\midrule 
WebSight VLM-8B & 55.9 & 86.6 & 77.3 & 79.4 & 87.6 \\
\texttt{\modelname}-18B & 78.5 & 96.4 & 74.3 & 67.0 & 85.8\\

\bottomrule
\end{tabular}
\endgroup
\caption{
Automatic evaluation results of the four fine-grained similarity measures and the high-level visual similarity with CLIP. The best result per dimension is highlighted in bold. 
}
\label{tab:auto_eval}
\end{table}
the most relevant and accessible baselines for our benchmark.
We observe that (1) GPT-4o is the best on all dimensions, while LLaVA 1.6-7B leads open-source models without specific finetuning. With specific finetuning, the performance of open-source models can approach that of commercial models like Gemini 1.0 Pro Vision.
(2) Text-augmented prompting successfully increases the block-match score and text similarity score on most tested models, especially those that are suboptimal in terms of text recognition, indicating the usefulness of providing extracted text elements.
(3) Self-revision has some minor improvement on block-match and position similarity for GPT-4V and Claude 3, but brings no improvement on Gemini Pro Vision and all other open-source models, potentially due to the limited capabilities of LLMs to do self-revision~\citep{Huang2023LargeLM}.
We provide an in-depth analysis of the learning process of our finetuned model in Section \ref{sec:learning process}.

\subsection{Human Evaluation}

While the above automatic metrics provide a fine-grained breakdown of model performance, it is also crucial to ask what users, the ultimate audience of these webpages, think of the generated webpages.
By recruiting human annotators (paid at the rate of \$16/hour) from Prolific, 
we conducted a series of human evaluations to compare across models and methods, as well as to assess the quality of AI-generated webpages directly. We sample 100 examples from our benchmark for the human evaluations. In all human evaluations, each question is annotated by $5$ human annotators, and we derive the results by majority voting. We provide all instructions that we provided to annotators in Appendix~\ref{appendix_human_annotation} and we outline the main protocols and results below.

\noindent
\textbf{Pairwise Model Comparison}
Following the conventional practice of evaluating instruction-following LLMs (e.g., \citep{lima,Dubois2023AlpacaFarmAS}), we ask human annotators to rank a pair of generated webpages (one from the baseline, the other from the tested method) to decide which one is more similar to the reference. 
We use Gemini Pro Vision Direct Prompting as the baseline 
and collect the other seven methods' Win/Tie/Lose rates against this baseline (we randomly shuffle the ordering to avoid position biases). Each pair will count as Win (Lose) only when Win (Lose) receives the majority vote ($\geq 3$). All other cases are considered Tie.

\begin{figure*}[t]
\centering
\captionsetup{font=footnotesize}
\includegraphics[width=2\columnwidth]{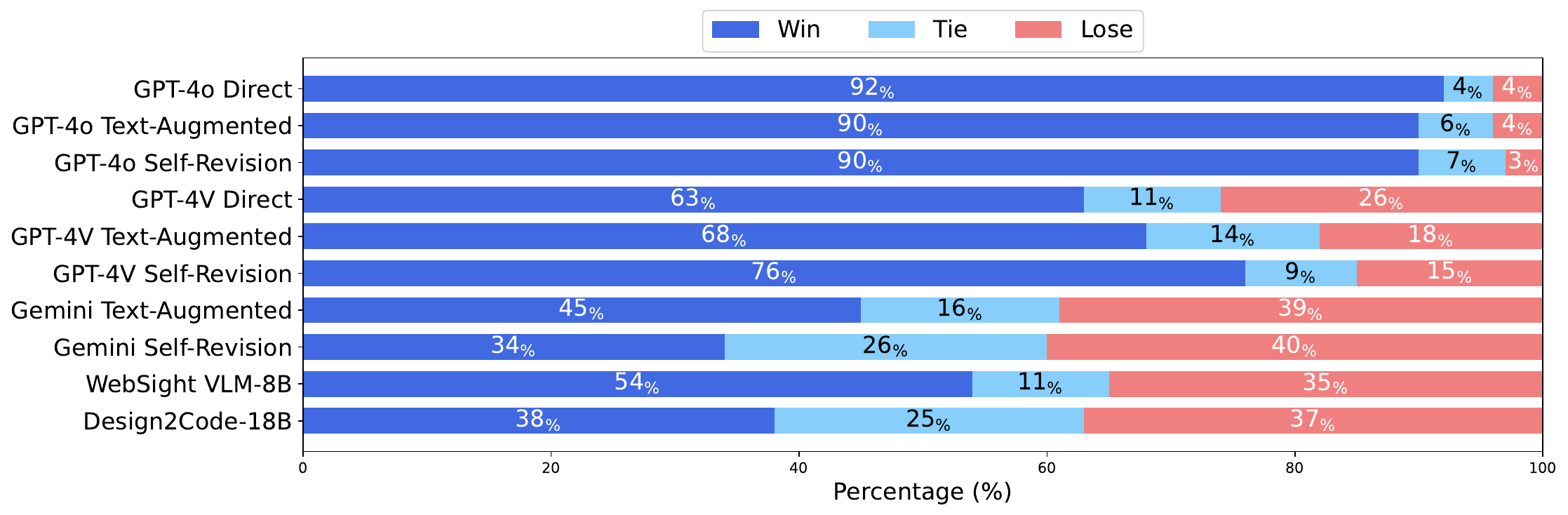}
\caption{Human pairwise preference evaluation results with Gemini Pro Vision Direct Prompting as the baseline (this method itself is not shown in the table since it serves as the baseline for pairwise comparison). 
We sample 100 examples, ask 5 annotators for each pair of comparisons, and take the majority vote on each example.
A higher win rate and lower loss rate suggest better quality as judged by human annotators.
}
\label{fig: human_eval_win_rate}
\end{figure*}

Based on the human evaluation in Figure \ref{fig: human_eval_win_rate}, we find that:
(1) GPT-4o is substantially better than other baselines, while neither text-augmented prompting nor self-revision prompting brings substantial improvement, probably due to the fact that direct prompting already correctly generates most of the text content and layout.
(2) GPT-4V is the second strongest model, while both text-augmented prompting and self-revision prompting can further improve over direct prompting. 
(3) Text-augmented prompting can slightly improve the Gemini direct prompting baseline, but further adding self-revision is not helpful. Intuitively, self-revision needs the model to understand the differences between the two given images (the reference screenshot and the screenshot of the initial model generation) and reflect them correspondingly in the modified HTML code, which is harder than leveraging text augmentation and thus might require more advanced model capabilities. 
(4) WebSight VLM-8B and our model \texttt{\modelname}-18B match Gemini direct prompting, suggesting that finetuning on a large amount of data can match commercial models in specific domains.

\noindent
\textbf{Direct Assessment} 
While the automatic and human evaluation offer a comparison among different models and methods, readers might still wonder how good are the best-performing models on this task. 
To offer a more intuitive answer to this question, we further ask human annotators to compare each reference webpage with the AI-generated webpage (using GPT-4V self-revision prompting) and directly assess the quality of the generated webpage. Concretely, we perform direct assessment from two perspectives (all examples are annotated by 5 annotators, and we take the majority vote; full instructions given to the annotators can be found in Appendix~\ref{appendix_human_annotation}): (I) Can the AI-generated webpage replace the original webpage? We shuffle the ordering of all examples
and ask annotators to judge whether the two webpages are similar enough in terms of appearance and content so that they can be deployed interchangeably. We find that 49\% of the AI-generated webpages are considered exchangeable with the reference webpages. (II) Is the reference webpage or AI generation better? We then ask a different question, where we shuffle the example ordering and ask annotators which webpage is better designed (annotators do not know which one is the reference and which one is AI-generated). 
Perhaps surprisingly, webpages generated by GPT-4V are preferred in 64\% cases, i.e., they are considered better designed than even the original reference webpages. 
We hypothesize it is possible that the model has more access to modern and popular webpage design principles~\citep{ivory2005evolution, beaird2020principles}, such that it can automatically improve the original design based on these best practices. 
This also opens up many new opportunities for future work on website design improvement tools. We provide some case study examples in Appendix~\ref{sec: qualitative Analysis}.

\begin{table}[t]
\centering
\captionsetup{font=footnotesize}
\footnotesize
\centering
\begingroup
\setlength{\tabcolsep}{2pt} 
\renewcommand{\arraystretch}{1.0} 
\begin{tabular}{lrrr}
\toprule
                  & \textbf{coef} & \textbf{std err} & \textbf{p} \\ \midrule
\textbf{Block-Match}     & 0.7429        & 0.142            & 0.000      \\
\textbf{Text}     & -0.3541        & 0.153            & 0.021      \\
\textbf{Position} & 0.7605         & 0.139            & 0.000      \\
\textbf{Color}    & 0.3461        & 0.107            & 0.001      \\
\textbf{CLIP}     & 0.4929       & 0.134            & 0.000      \\ \bottomrule
\end{tabular}
\endgroup
\caption{
Results on predicting human annotations (Win/Lose) via logistic regression using different metrics as features.
}
\label{tab:coe analysis}
\end{table}

\begin{table}[t]
\centering
\captionsetup{font=footnotesize}
\footnotesize
\centering
\begingroup
\setlength{\tabcolsep}{2pt} 
\renewcommand{\arraystretch}{1.0} 
\begin{tabular}{l|cccc|c}
\toprule
 & Block & Text & Position & Color & CLIP  \\ 
\midrule
\multicolumn{6}{c}{GPT-4o} \\
\midrule 
Direct         & 56.6 & 89.8 & 78.6 & 81.9 & 87.1\\ 
Text-Augmented & 67.7 & 95.2 & 77.5 & 81.5 & 87.5\\ 
Self-Revision & 72.1 & 96.4 & 81.1 & 82.4 & 88.2\\
\midrule
\multicolumn{6}{c}{GPT-4o Mini} \\
\midrule 
Direct         & 57.7 & 90.7 & 77.9 & 77.5 & 86.3\\ 
Text-Augmented & 69.5 & 97.0 & 77.9 & 79.1 & 86.1\\ 
Self-Revision & 70.3 & 96.9 & 77.9 & 78.6 & 86.0\\
\midrule
\multicolumn{6}{c}{Claude 3.5 Sonnet} \\
\midrule 
Direct         & 61.7 & 91.1 & 83.0 & 84.4 & \textbf{89.5}\\ 
Text-Augmented & \textbf{75.1} & 97.6 & \textbf{83.4} & \textbf{84.9} & 89.0\\ 
Self-Revision & 71.9 & 96.5 & 82.6 & 83.0 & 88.8\\
\midrule
\multicolumn{6}{c}{Claude 3 Opus} \\
\midrule 
Direct         & 57.1 & 88.7 & 74.2 & 72.4 & 85.8\\ 
Text-Augmented & 73.6 & 97.0 & 75.6 & 72.6 & 85.7\\ 
Self-Revision & 73.3 & 95.9 & 76.6 & 70.0 & 85.6\\
\midrule
\multicolumn{6}{c}{Gemini 1.5 Pro} \\
\midrule 
Direct         & 72.3 & 95.4 & 80.9 & 80.5 & 87.5\\ 
Text-Augmented & 73.7 & 95.9 & 79.8 & 79.1 & 88.2\\ 
Self-Revision & 71.2 & 96.6 & 80.9 & 78.4 & 87.9\\
\midrule 
\multicolumn{6}{c}{Gemini 1.5 Flash} \\
\midrule 
Direct         & 69.2 & 96.6 & 78.0 & 80.2 & 87.3\\ 
Text-Augmented & 72.7 & 97.4 & 79.4 & 78.2 & 87.6\\ 
Self-Revision & 72.2 & \textbf{97.5} & 79.4 & 77.9 & 87.6\\
\bottomrule
\end{tabular}
\endgroup
\caption{
Evaluation results for \texttt{\modelname-HARD}. We show the results for flagship commercial models.
}
\label{tab:new_eval}
\end{table}

\begin{figure*}[t]
  \centering
  \includegraphics[trim={1cm 1cm 1cm 1cm},width=2\columnwidth]{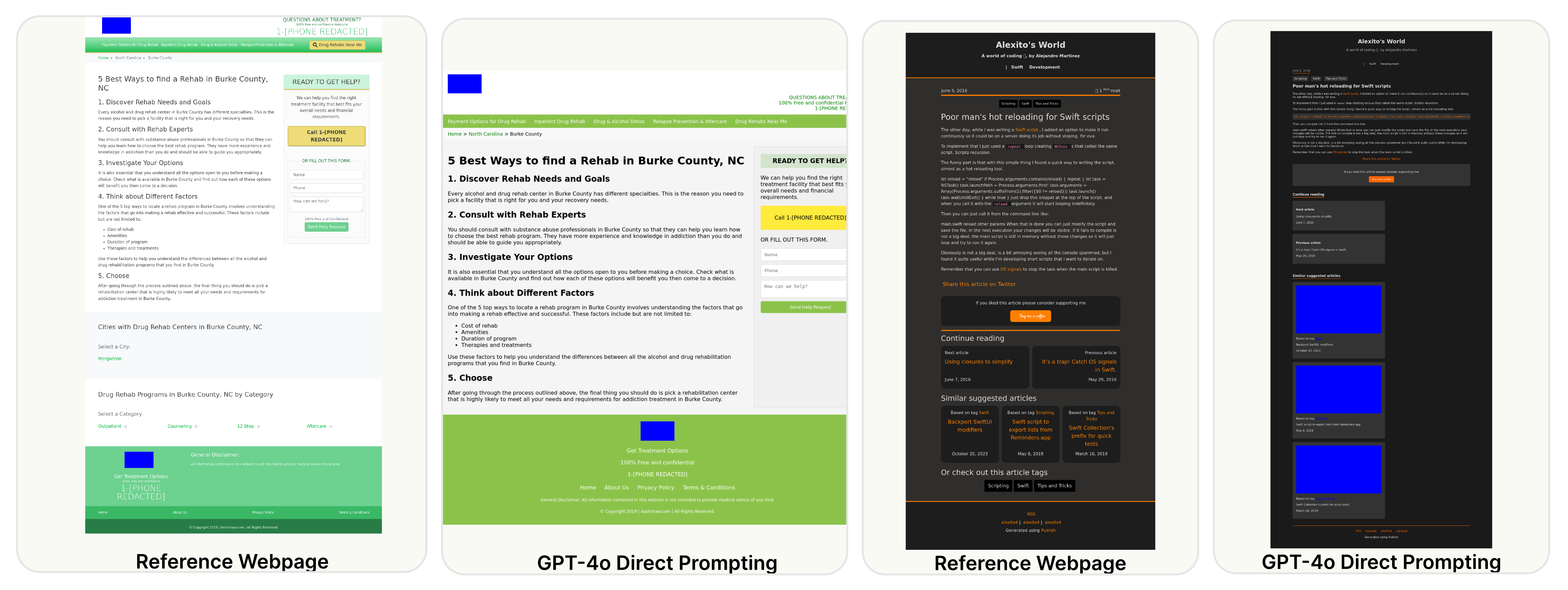}
  \caption{Example of GPT-4o direct prompting.}
  \label{fig:gpt4o}
\end{figure*}

\begin{figure*}[t]
  \centering
  \includegraphics[trim={1cm 1cm 1cm 1cm},width=2\columnwidth]{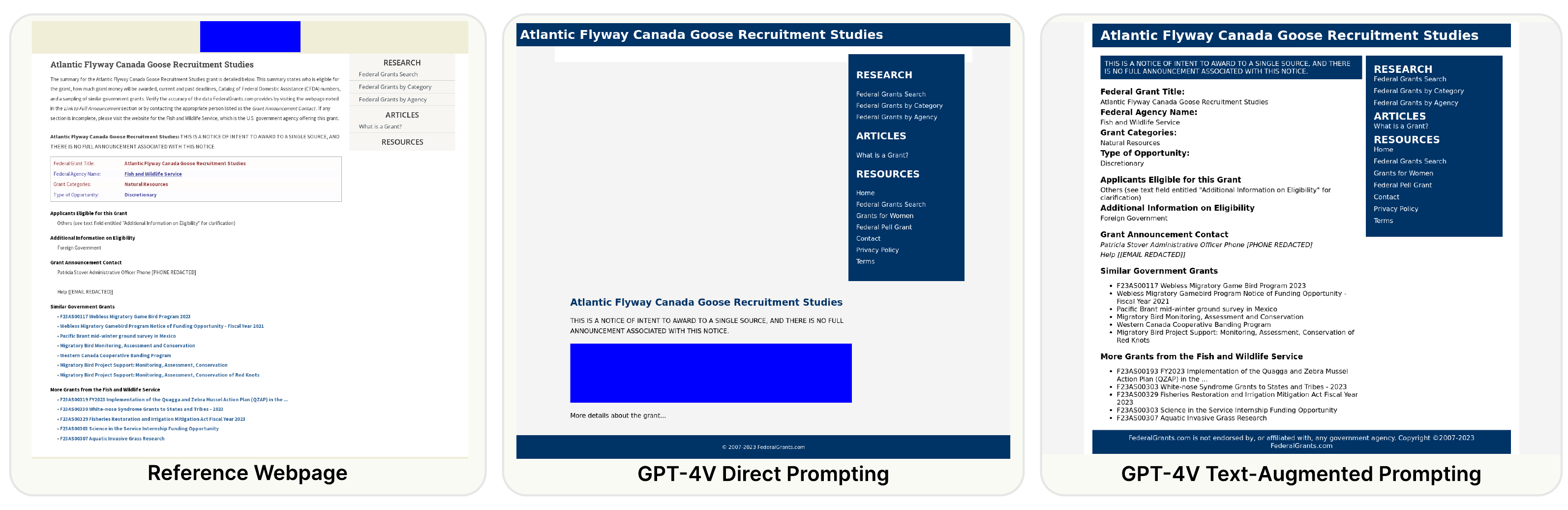}
  \caption{Example of text-augmented prompting improving over the direct prompting baseline, where missing texts are successfully generated.}
  \label{fig:text_augmentation_prompting}
\end{figure*}

\begin{figure*}[t]
  \centering
  \includegraphics[trim={1cm 1cm 1cm 1cm},width=2\columnwidth]{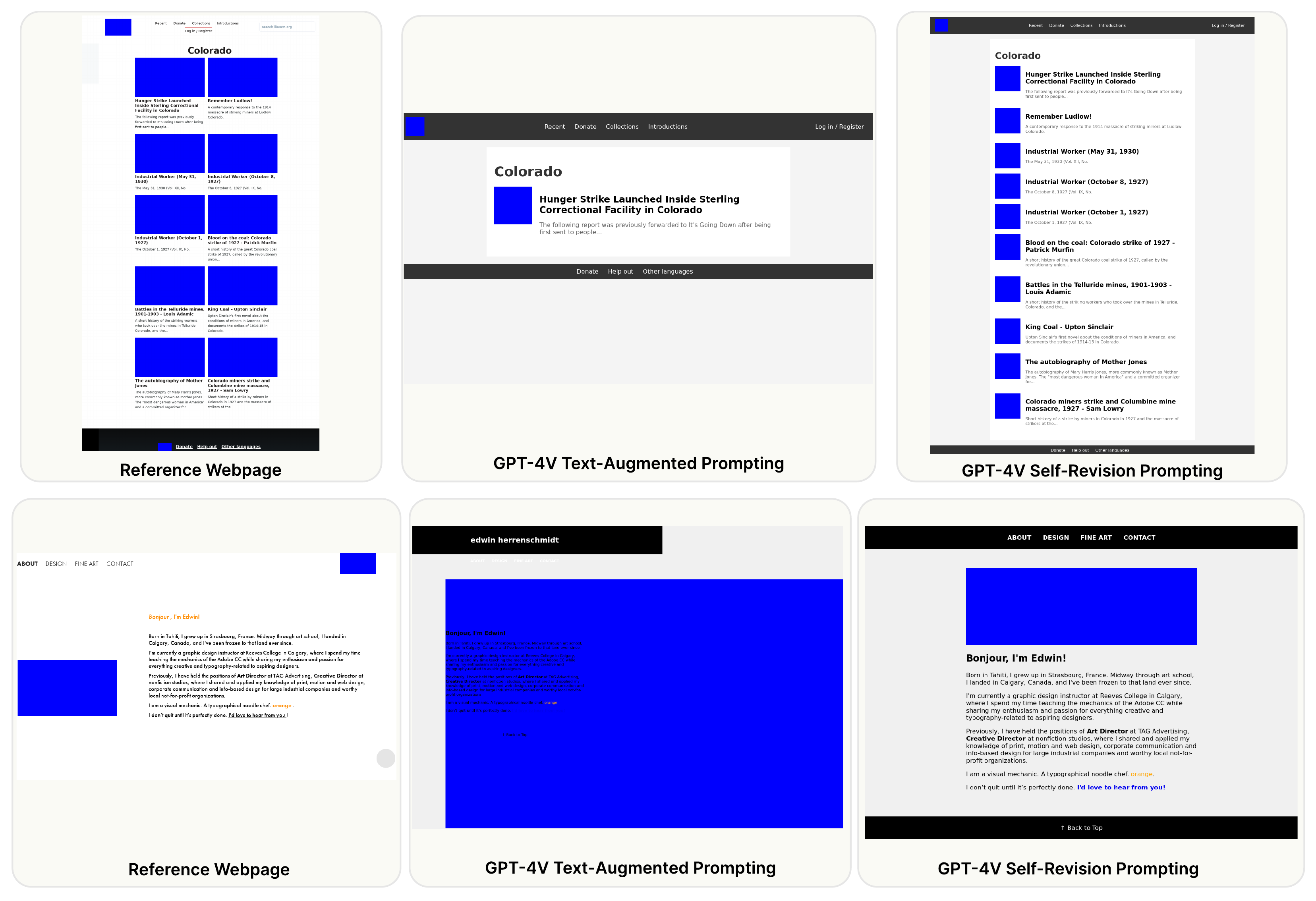}
  \caption{Examples of self-revision prompting improving over text-augmented prompting. The self-revision can either add the missing texts or fix layout errors. 
  }
  \label{fig:self_revision_prompting}
\end{figure*}


\subsection{Automatic Evaluation vs Human Evaluation}
It is worth noting that there are 
some interesting discrepancies between the automatic evaluation results and human evaluation results. For example, human evaluation ranks GPT-4V self-revision prompting better than text-augmented prompting, while the automatic metrics show mixed results. Moreover, even though humans rank WebSight VLM-8B as better than \texttt{Design2Code-18B}, it has much worse block-match and text similarity as measured by the automatic metrics. In this part, we take a closer look at such discrepancy and discuss 
why such discrepancy is a feature rather than a bug. 

\textbf{Human annotation replies only on high-level features.} We study whether we can predict human pairwise preferences using automatic metrics. Specifically, we randomly split the 100 annotated examples into a 50\% training set and a 50\% test set, leading to 435 pairwise human annotations (we only consider Win/Lose) for both training and testing. Given one reference $R$ and two candidates $G_1, G_2$, we use the difference of each dimension (e.g., $\mathbf{match_{block}}(R, G_1) - \mathbf{match_{block}}(R, G_2)$) as features and predict Win ($1$) or Lose ($0$) by logistic regression \footnote{We normalize the feature before calculating the differences and add a constant term before logistic regression.}. The derived logistic regression model achieves 79.9\% accuracy on the test set, and the features' coefficients and significance are in Table \ref{tab:coe analysis}. Interestingly, we find that text similarity has a negative and least significant association with human judgment. In contrast, all other similarity measures show positive and significant associations with human judgment.
This suggests that humans usually pay more attention to high-level visual effects like layouts, colors, and the existence of contents rather than the detailed content, reflecting the top-down processing \citep{gilbert2013top} of humans. 
We argue that human evaluation should not be blindly trusted as the oracle here due to their cognitive bias to only consider ``principle components'' of the webpages. Instead, both high-level similarity (human pairwise preference and CLIP similarity) and low-level elements (fine-trained block-wise similarity) should be considered when evaluating new models and methods.

\vspace{-0.01in}
\subsection{Case Study}
\vspace{-0.01in}
Figure~\ref{fig:gpt4o} shows examples from GPT-4o, which generates similar layouts and color styles without prompting techniques. For weaker models like GPT-4V, text-augmented prompting improves recall, especially for texts, as seen in Figure~\ref{fig:text_augmentation_prompting}, where it raises the block-match score from 0.25 to 0.84. We then analyze self-revision’s impact on text-augmented prompting. In Figure~\ref{fig:self_revision_prompting}, self-revision recovers missing webpage elements, increasing the block-match score from 0.48 to 1.00 and CLIP similarity from 0.87 to 0.91. It also corrects layout errors, boosting CLIP similarity from 0.85 to 0.91.

\vspace{-0.01in}
\subsection{\texttt{\modelname-HARD}}
\vspace{-0.01in}

To understand what makes a webpage difficult to generate, we compute the correlation between automatic metrics and various difficulty indicators, including (1) the total number of tags in the reference implementation, (2) the number of unique tags in the reference implementation, and (3) DOM tree depth of the reference implementation. Appendix Table~\ref{tab:difficulty_correlation} shows that the total number of tags is a strong indicator of difficulty. We also find that non-English webpages are usually harder to generate, probably due to the limited pretraining data.

To this end, we introduce a more difficult version of our benchmark using the same filtering process, named \textbf{\texttt{\modelname-HARD}}, containing 80 hard examples with unique challenges like being extremely long (26\% examples have more than 500 HTML tags) and non-English content (19\% examples). In table \ref{tab:new_eval}, we observe a significant performance drop on GPT-4o compared to the ``easy'' version of our dataset, and state-of-the-art models fail to generate 30\% - 40\% of the block elements (Block-Match). Interestingly, GPT-4o can steadily improve performance through text augmentation and self-revision, while Gemini 1.5 Pro cannot.

\section{Related Work}


\noindent
\textbf{UI Automation}
\citet{nguyen2015reverse} reverse engineer mobile UI by identifying elements through classic text recognition and computer vision techniques (OCR, edge detection, etc).
Pix2Code \citep{beltramelli2018pix2code} builds an end-to-end system for UI-to-code transformation based on CNN and RNN, which cannot deal with complex visual encoding and long text decoding. \citet{Robinson2019Sketch2codeGA, mockup2html} further incorporate neural network-based object detection and semantic segmentation into the pipeline. Recently, \citet{Soselia2023LearningUR} utilize more advanced visual encoders (e.g., ViT, \citealp[]{vit}) and language decoders (e.g., LLaMA, \citealp[]{llama1, llama2}) and finetune the pipeline using visual similarity. ~\citet{Jiang2024Graph4GUIGN} use Graph Neural Networks to represent UI elements and designers
by autocompleting partially completed GUI designs.
We advance this thread by offering the first UI automation benchmark with real-world webpages and benchmarking state-of-the-art MLLMs. 

\noindent
\textbf{Code LLMs and Programming Support Tools} Our work also connects to code language models and programming support tools. LLMs trained on code, such as Codex~\citep{Chen2021EvaluatingLL}, StarCoder~\cite{StarCoder}, InCoder~\citep{Fried2022InCoderAG}, CodeLlama~\citep{Rozire2023CodeLO}, and DeepSeek-Coder~\citep{deepseekcoder}, enable a wave of programming support applications such as automatic code completion and infilling, and allowing users to chat with a codebase.  This also leads to a new wave of HCI studies on designing better programming tools to facilitate human-AI collaboration~\citep{copilot_productivity,Vasconcelos2023GenerationPA,Liang2023ALS}. Our benchmark offers realistic evaluation for code LLMs and aims to enable more powerful programming support to front-end designers who do not have to code by themselves and can just collaborate with LLMs.  

\section{Conclusion}

This work introduced the \texttt{Design2Code(-HARD)} benchmark consisting of diverse and challenging real-world webpages as test examples. We developed comprehensive automatic metrics and conducted human evaluations to compare various multimodal LLMs. We showed that state-of-the-art models can generate well-formed websites on some easy examples but still struggle with more complex examples, while open-source models are far behind commercial models, leaving ample room for future improvement. For future work, a promising direction is to generate UIs that support dynamic user interactions. 

\section*{Limitations}
\label{sec:limitations}

We believe \modelname can serve as a useful benchmark to power many future research directions. We highlight a few current limitations of \modelname and how future work could address them: 

\begin{enumerate}
    \item Better prompting techniques for multimodal LLMs, especially in handling complex webpages, for example, by incrementally generating different parts of the webpage. 

    \item Our preliminary experiments showed the difficulty of directly training on real webpages since they are too long and noisy, future work could explore data cleaning pipelines to make such training stable. 

    \item Extending beyond screenshot-only inputs, for example, to collect Figma frames or sketch designs from front-end designers as the test input. Such extension also requires careful re-design of the evaluation paradigm. 

    \item Extending from static webpages to also include dynamic webpages. This also requires the evaluation to consider interactive functions beyond just visual similarity. 
\end{enumerate}

\section*{Ethical Considerations}
\label{sec:ethics}

\paragraph{Privacy} We used the dataset C4 which is released under ODC-By license, allowing free share, modification, and use subject to the attribution requirements. We release our dataset under the same license. Moreover, when performing manual filtering, we explicitly filtered out webpages containing private or sensitive information (e.g., dating website profiles).

\paragraph{Dual Use} Despite our intention of democratizing webpage building, we recognize the potential dual use danger of \modelname technologies, such as automated generation of malicious websites, or even generating code for licensed websites. 
We emphasize is intended for research purposes and for the community to better understand multimodal LLM capabilities. 
We will provide clear ethical use guidelines for all data, code, and model releases to define acceptable and unacceptable use cases. 

\section*{Acknowledgement}
We thank Aryaman Arora, Jihyeon Je, Irena Gao, Will Held, Ryan Louie, Weiyan Shi, Yutong Zhang, Dora Zhao, Rose Wang, Caleb Ziems, Michael Ryan, Camille Harris, Harshit Joshi, Yijia Shao, Jiaao Chen, Omar Shaikh, Julie Kallini, Lucia Zheng, Julia Kruk, Yanchen Liu, Tianyu Gao and Tristan Thrush for their helpful comments and discussion. 
This work is supported in part by a grant from Google and ONR to DY.

\bibliography{custom}

\newpage

\appendix

\section{Additional Dataset Statistics}
\label{appendix_dataset_statistics}

\begin{table*}[h]
\small
\centering
\begin{tabular}{l|c|c|c}
\toprule
& \textbf{WebSight (Huggingface)} & \textbf{\modelname (Ours)} & \textbf{\modelname-HARD (Ours)} \\ 
\midrule
Purpose & Training & Testing & Testing \\ 
Source & Synthetic (Deepseek-Coder) & Real-World (C4) & Real-World (GitHub) \\ 
Size & 823K & 484 & 80 \\
Avg Tag Count & 19{\small $\pm 8$} & 158{\small $\pm 100$} & 251{\small $\pm 232$} \\
Avg DOM Depth & 5{\small $\pm 1$} & 13{\small $\pm 5$} & 10{\small $\pm 4$} \\
Avg Unique Tags & 10{\small $\pm 3$} & 22{\small $\pm 6$} & 22{\small $\pm 5$} \\ 
\bottomrule
\end{tabular}
\vspace{0.1in}
\caption{
Comparison of datasets statistics between the WebSight dataset and our new \modelname benchmark. WebSight only provides the training set while \modelname only provides the test set. Examples in our \modelname benchmark are much more complex on all measures and have a wider variety of difficulty levels as indicated by the bigger standard deviations. 
}
\label{tab:dataset_comparison}
\end{table*}

We present the table of most frequent HTML tags in Table~\ref{tab:frequency_tags}.

\begin{table*}[t]
\centering
\begin{tabular}{lc|lc|lc}
\toprule
Tag    &  Frequency & Tag & Frequency  & Tag & Frequency  \\ 
        \midrule
\textbf{<div>}    &   17790  & \textbf{<style>} & 1181  & \textbf{<head>} & 486 \\
\textbf{<a>}    &   13309  & \textbf{<td>} & 997 & \textbf{<body>} & 486 \\
\textbf{<li>} &  6883  & \textbf{<input>} &  995 & \textbf{<tr>} & 436 \\
\textbf{<span>}   &  6813  & \textbf{<h3>} & 759 & \textbf{<b>} & 429  \\
\textbf{<meta>}    &  4629  & \textbf{<h2>} & 709 & \textbf{<nav>} & 416 \\ 
\textbf{<p>}   &   3413 & \textbf{<strong>} &  595  & \textbf{<i>} & 400 \\ 
\textbf{<br>}   &  2453  & \textbf{<h1>} & 536  & \textbf{<section>} & 381  \\
\textbf{<ul>}   & 2078  & \textbf{<button>} & 525  & \textbf{<label>} &  339 \\ 
\textbf{<img>}   &  1870  & \textbf{<title>} & 492  & \textbf{<form>} &  292 \\ 
\textbf{<option>}   & 1194  & \textbf{<html>} & 486  & \textbf{<h4>} &  289 \\ 
\bottomrule
\end{tabular}
\vspace{0.1in}
\caption{
The most frequent HTML tags in the reference implementations of our benchmark examples. 
}
\label{tab:frequency_tags}
\end{table*}

\begin{table*}[t]
\centering
\begin{tabular}{lrrr}
\toprule
                  & \textbf{coef} & \textbf{std err} & \textbf{p} \\ \midrule
\textbf{const}     & 0.5540        & 0.139            & 0.000      \\
\textbf{Block-Match}     & 0.6238  & 0.131            & 0.000      \\
\textbf{Position} & 0.7504         & 0.141            & 0.000      \\
\textbf{Color}    & 0.3443        & 0.107            & 0.001      \\
\textbf{CLIP}     & 0.4630       & 0.132            & 0.000      \\ \bottomrule
\end{tabular}
\vspace{0.1in}
\caption{
Coefficients for the learned linear regression model to simulate win rate.
}
\label{tab:coe analysis without text}
\end{table*}

\begin{table*}[t]
\centering
\begin{tabular}{lc|lc|lc}
\toprule
\multicolumn{2}{c}{Total Num of Tags}   &  \multicolumn{2}{c}{Num of Unique Tags} & \multicolumn{2}{c}{DOM Tree Depth}  \\ 
        \midrule
Metric & Corr & Metric & Corr & Metric & Corr \\
\midrule 
Block-Match & -0.28* & Block-Match & -0.16* & Block-Match  & -0.04 \\ 
Text & -0.13* & Text & -0.08 & Text  & 0.01 \\ 
Position & -0.19* & Position & -0.15* & Position & -0.10*  \\ 
Color & -0.13* & Color & -0.09 & Color  & -0.04 \\ 
CLIP & -0.12 & CLIP & -0.02 & CLIP  & 0.03 \\ 
\bottomrule
\end{tabular}
\vspace{0.1in}
\caption{
Correlation between automatic metrics and three proxy difficulty indicator variables on GPT-4V self-revision prompting. The total number of tags is the strongest indicator, where webpages with more tags tend to be more challenging for the model. * indicates p-value $<0.05$. 
}
\label{tab:difficulty_correlation}
\end{table*}

\section{Test Set Curation}
\label{sec: detailed test set curation}

Our overall goal is to obtain a set of well-formed webpages that represent diverse real-world use cases. We follow the following steps for automatic processing and manual filtering. 

\noindent
\textbf{Automatic Length and Layout Filtering} We first apply a round of automatic filtering. We strip all comments from the code files and then apply a length filter to exclude examples where the source code file has over 100k tokens (based on the \texttt{GPT-2} tokenizer), as a way to avoid excessively long webpages that current multimodal LLMs cannot process as input or cannot decode such long outputs.
Next, we filter all webpages whose layout consists of only images or only texts, in which cases the layout designs tend to be too simplistic to be interesting for benchmarking. This results in 14k webpages after filtering and deduplication. 

\noindent
\textbf{Making Webpages Stand-alone} We assume a setting where we will only provide the screenshot of the webpage for the model, without providing all the external dependencies such as multimedia files (images, audio, videos, etc.). To make this possible, we strip all such external file dependencies to make all the webpages stand-alone, this includes: removing all \texttt{<script><audio><iframe><map><svg>} tags, removing all \texttt{<link>} tags that link to external sites, removing all href links in \texttt{<a>} tags, and removing all external files in \texttt{<object>} elements. For all the image and video files, we replace them with a placeholder file, and during benchmarking we will instruct the models to insert this placeholder file wherever applicable to preserve the original layout. 

\noindent
\textbf{Manual Curation} After the above processing, we perform a final round of manual curation to filter examples based on the following criteria: (1) The webpage has no external file dependency and can render in a stand-alone manner from the processed code file and provided placeholder image file. (2) The webpage does not contain any private, sensitive, or potentially harmful information (e.g., we removed profile pages from dating websites). (3) The rendered webpage is well-formatted (e.g., there should not be overlaps between different layout elements and the automatic processing above should not disrupt any part of the webpage design). The first two authors of this paper performed this curation step by checking every single example from the sampled 7k examples. They first annotated 200 examples together to reach an $75\%$ agreement, then split the annotation work on 7k randomly sampled examples from the filtered set of 14k examples above. This entire manual curation process took approximately one week. We try to keep the selected high-quality webpages as diverse as possible while adding new ones. In the end, we obtained 484 test examples that we use as our benchmark. 

\section{Text Detection and Merging Details}
\label{sec:detection}

The common approach to detect the texts in a given screenshot is to use OCR tools \citep{OCRReview}, which returns a list of text segments with their bounding boxes. However, in our case, we find that open-source OCR tools usually output noisy outputs, which may affect the stability of downstream evaluation. Since we already have the source HTML codes for reference webpage screenshots, we apply an alternative approach: we alter the color differently for different text segments in the source HTML code and detect text segments in the webpage by taking two extra screenshots and tracking pixels with different colors. This helps us locate text segments from the HTML source code in the screenshots without text recognition errors.

Based on the two sets of detected blocks, we use the Jonker-Volgenant algorithm \citep{matchingalgo} (implemented in Scipy \footnote{\url{https://docs.scipy.org/doc/scipy/reference/generated/scipy.optimize.linear_sum_assignment.html}}) to get the optimal matching $M$ between $R$ and $G$, where $(p, q) \in M$ indicates $r_p$ is matched with $g_q$. Specifically, we use the negative sequence similarity between textual contents ($-\mathbf{sim_{text}}(,)$) to initialize the cost matrix and ignore the matched pairs with a sequence similarity lower than $0.5$. Since detected text blocks might be in different granularity, we also enumerate merging neighbor text blocks to search for matching with the highest similarity. However, the matching may still not be perfect, especially when there are large granularity differences (our search does not consider merging non-contiguous blocks).

\section{Prompting details}
\label{sec:prompts}

We use the following prompt for direct prompting:

\begin{small}
\texttt{You are an expert web developer who specializes in HTML and CSS. A user will provide you with a screenshot of a webpage. You need to return a single html file that uses HTML and CSS to reproduce the given website. Include all CSS code in the HTML file itself. If it involves any images, use "rick.jpg" as the placeholder. Some images on the webpage are replaced with a blue rectangle as the placeholder, use "rick.jpg" for those as well. Do not hallucinate any dependencies to external files. You do not need to include JavaScript scripts for dynamic interactions. Pay attention to things like size, text, position, and color of all the elements, as well as the overall layout. Respond with the content of the HTML+CSS file.}
\end{small}

We use the following prompt for self-revision prompting:
\begin{small}
\texttt{You are an expert web developer who specializes in HTML and CSS. I have an HTML file for implementing a webpage but it has some missing or wrong elements that are different from the original webpage. The current implementation I have is: [generated code from text-augmented prompting]. I will provide the reference webpage that I want to build as well as the rendered webpage of the current implementation. I also provide you all the texts that I want to include in the webpage here: [extracted texts from the original webpage]. Please compare the two webpages and refer to the provided text elements to be included, and revise the original HTML implementation to make it look exactly like the reference webpage. Make sure the code is syntactically correct and can render into a well-formed webpage. You can use "rick.jpg" as the placeholder image file. Pay attention to things like size, text, position, and color of all the elements, as well as the overall layout. Respond directly with the content of the new revised and improved HTML file without any extra explanations.}
\end{small}

Details of tested models: GPT-4o: \texttt{gpt-4o-2024-05-13}, GPT-4V: \texttt{gpt-4-1106-vision-preview}, Claude 3 Opus: \texttt{claude-3-opus-20240229}, Gemini 1.0 Pro Vision \texttt{gemini-1.0-pro-vision} \footnote{We keep getting empty responses with unexplained errors while testing Gemini 1.5.}. For all commercial models, we use greedy decoding and set maximum new tokens to 4096.

For open-source models, we found that they tend to generate repetitive content for HTML/CSS in our preliminary experiments. We use a temperature of $0.5$ and a repetition penalty of $1.1$ during sampling, the same as testing the finetuned models. Note that other open-source models like Qwen-VL-Chat \citep{qwenvl} and Mantis \citep{jiang2024mantis} fail to generate HTML format in more than 80\% cases, which are not reported here. Since most of the open-source models only support single-image input, we concatenate the reference screenshot and the generated screenshot into one image before prompting, following \citet{jiang2024mantis}.

\section{Finetuning details of \texttt{\modelname}-18B}
\label{sec: finetuning details}

We use CogAgent-18B~\cite{cogagent} as our base model, which supports high-resolution input ($1120 \times 1120$) and is pretrained on intensive text-image pairs \citep{kakaobrain2022coyo-700m, schuhmann2022laion5b}, synthetic documents \citep{kim2022ocr}, LaTeX papers \citep{blecher2023nougat}, and a small amount of website data. We then finetune the base model with the WebSight dataset. While the original WebSight dataset has 823K examples, we only randomly sample 20\% for training due to the limited computation resources. We also reverse the order of HTML style and body as we find that it leads to a lower loss in our preliminary experiment. Note that we have also experimented with training on real-world webpage data scraped from the C4 training set. Such training is extremely unstable and difficult because real-world code implementation data tend to be extremely long and noisy, resulting in even lower performance than training on synthetic data. We thus leave such exploration to future work. Specifically, We use LoRA \citep{lora} to fine-tune the base model, where the LoRA modules are added to the language decoder with LoRA rank $8$. Using a batch size of $32$ and a learning rate of 1e-5, we fine-tune the model for $5000$ steps with $100$ steps warmup. Using $4 \times$ NVIDIA A6000, this takes about 2 days of training. We use a temperature of $0.5$ and a repetition penalty of $1.1$ during inference and select the best checkpoint based on the average of all automatic metrics on a small dev set (20 examples).

\section{Human Annotation Details}
\label{appendix_human_annotation}

We restrict the annotators to people in the U.S. who have completed 2,500 surveys with a pass rate of 98\% or higher. In total, there are 60 participants.
In the instructions, the annotators are asked to check the pair following the order of priority (content $>$ layout $>$ style). This priority list is based on two intuitions: (i) Layout comparison is only meaningful when the content is (almost) complete. (ii) The style of independent elements is easier to fix than the layout of multiple elements. The detailed instructions are below:

\begin{center}
\begin{myquote}

\textbf{Task Overview}

In this survey, you will be given a reference webpage’s screenshot, as well as two candidate webpages (Example 1 and Example 2) that try to replicate the reference webpage. Your task is to judge which of the two candidates is closer to the reference.

Each (Reference, Example 1, Example 2) is presented in a row, where the original boundary of screenshot is marked by black.

\textbf{Comparison Guide}

\textbf{Initial Step: Content Check}
\begin{itemize}
    \item \textbf{Text Content:} Examine if the text on the candidate webpages matches the reference. Pay special attention to missing or extra content, especially key elements like titles.
    \item \textbf{Image Content:} Assess the placement of the blue placeholder blocks (for images).
    \item \textbf{Primary Judgment Criterion:} If one example has significant missing or additional content compared to the other, it should be considered less similar to the reference.
\end{itemize}

\textbf{Second Step: Layout Check}
\begin{itemize}
    \item \textbf{Element Arrangement:} If the content (text and images) of both examples is similarly good or bad, proceed to evaluate the arrangement of these elements. Check if their organization, order, and hierarchy match the reference.
    \item \textbf{Secondary Judgment Criterion:} If differences in layout are observed, the example with the layout most similar to the reference should be rated higher.
\end{itemize}

\textbf{Final Step: Style Check}
\begin{itemize}
    \item \textbf{Style Attributes:} Only if Example 1 and Example 2 are comparable in content and layout, examine the style elements like font style, color, and size.
    \item \textbf{Tertiary Judgment Criterion:} In cases where content and layout are equally matched, preference should be given to the example with style attributes closer to the reference.
\end{itemize}

\textbf{Overall Judgment}

Based on the criteria in the order of priority (Content > Layout > Style), make an overall judgment on which example (Example 1 or Example 2) is more similar to the reference webpage.

\textbf{Judgment Options}

1. Select "Example 1 better" if Example 1 is closer to the reference.

2. Select "Example 2 better" if Example 2 is closer to the reference.

3. Opt for "Tie" only if both examples are similarly accurate or equally distant from the reference.

\textbf{Additional Tips}

1. Use zoom-in for detailed inspection.

2. Focus on major discrepancies in each step before moving to the next.

3. Your judgment should be based on a cumulative assessment of content, layout, and style.

\end{myquote}
\end{center}

\begin{figure*}[h]
  \centering
  \includegraphics[width=2\columnwidth]{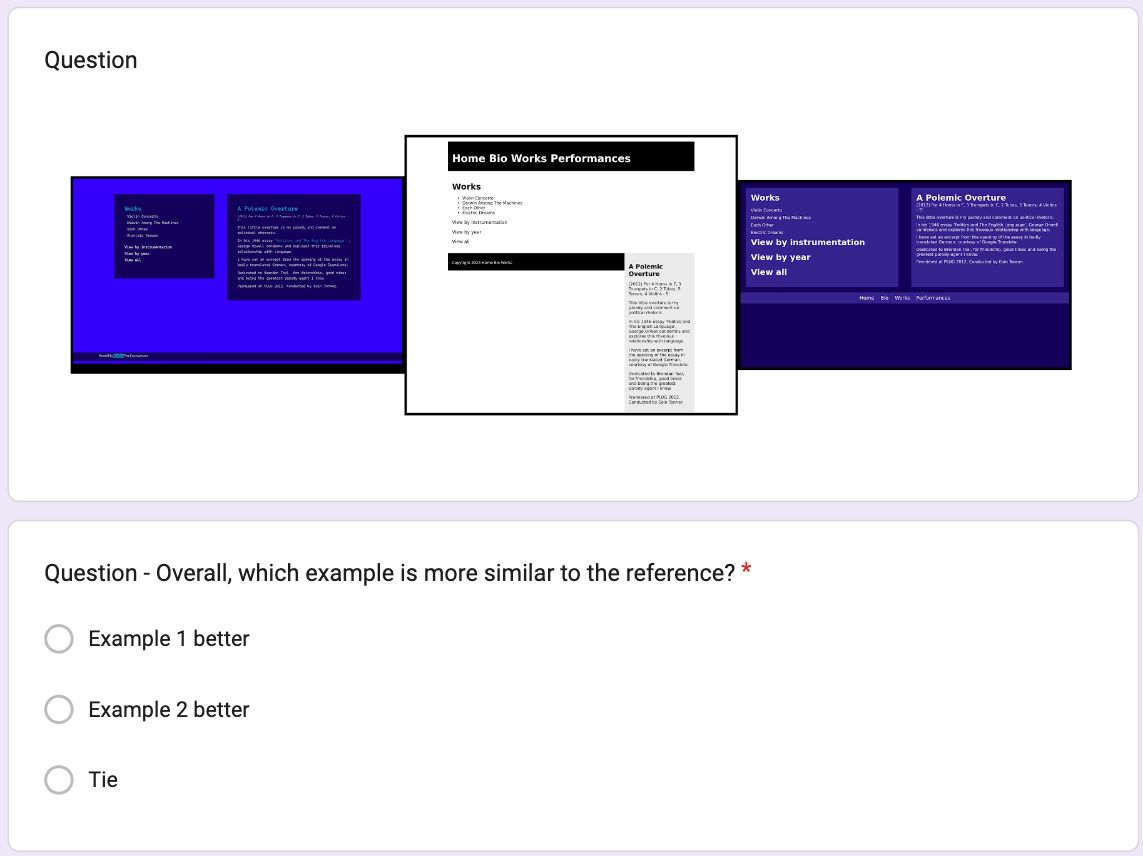}
  \caption{User Interface for pairwise model comparison.}
  \label{fig:UI human annotation 1}
\end{figure*}

We also provide 8 examples after the instruction. The UI of the annotation question is Figure \ref{fig:UI human annotation 1}. Fleiss' kappa for pairwise model comparison is $0.46$ ($5$ annotators).

Furthermore, we provide the instructions for direct assessment (comparing the reference and webpages generated by GPT-4V self-revision prompting). The Fleiss' kappa is $0.32$ ($5$ annotators) for the first question and $0.26$ ($5$ annotators) for the second question.

\textbf{Can the AI-generated webpage replace the original webpage?}

\begin{center}
\begin{myquote}

\textbf{Task Overview}

In each question, you will be given two webpage screenshots.

\textbf{By comparing the two webpages, you need to decide whether they are exchangeable.}

Please zoom in to take a closer look at the screenshots if necessary.

You should answer "Yes", if:

1. They look roughly similar.

2. They have similar content.

3. They can serve the same functions.

(Minor details don't matter that much)

Otherwise, you should answer "No".

\end{myquote}
\end{center}

\textbf{Is the reference webpage or AI generation better?}

\begin{center}
\begin{myquote}

\textbf{Task Overview}

In each question, you will be given two webpage screenshots.

\textbf{By comparing the two webpages, you need to decide which one is better.}

Please zoom in to take a closer look at the screenshots if necessary.

To decide which one is better, you might consider the following aspects:

1. More readable

2. Better layout

3. Better style

\end{myquote}
\end{center}

\begin{figure*}[t]
  \centering
  \includegraphics[width=1.4\columnwidth]{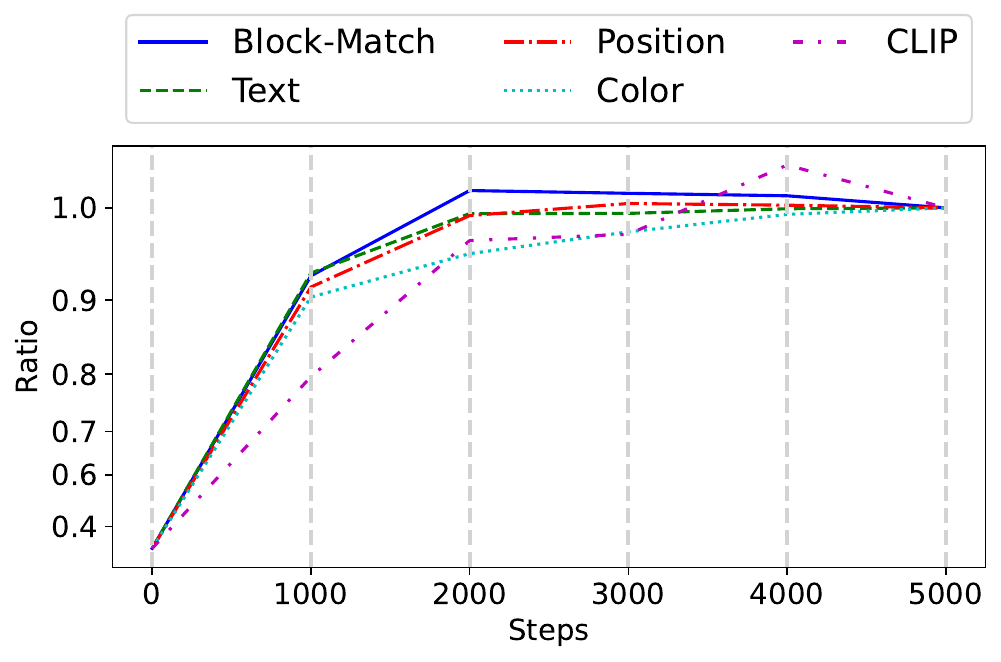}
  \caption{Learning process for different automatic evaluation dimensions, where we plot the performance for the base model checkpoint and all training checkpoints. For each dimension, the score is re-scaled so that it is $0$ before training ($0$ steps) and $1$ after training ($5000$ steps). The y-axis is rescaled to highlight the differences for bigger values.}
  \label{fig:skill_learning_ratio}
\end{figure*}

\begin{table}[ht]
\centering
\captionsetup{font=footnotesize}
\footnotesize
\centering
\begingroup
\setlength{\tabcolsep}{2pt} 
\renewcommand{\arraystretch}{1.0} 
\vspace{-3mm}
\begin{tabular}{lcc}
\toprule
\textbf{Model} & \textbf{Simulated} & \textbf{Annotated} \\ \midrule
GPT-4o Direct & 96.1 & 96.0 \\
GPT-4o Text-Augmented & 94.8 & 96.0 \\
GPT-4o Self-Revision & 95.0 & 97.0 \\
GPT-4V Direct & 81.0 & 74.0 \\
GPT-4V Text-Augmented & 81.4 & 82.0 \\
GPT-4V Self-Revision & 85.7 & 85.0 \\
Gemini Text-Augmented & 51.9 & 61.0 \\
Gemini Self-Revision & 50.4 & 60.0 \\
WebSight VLM-8B & 58.3 & 65.0 \\
\texttt{\modelname}-18B & 58.3 & 63.0 \\ \bottomrule
\end{tabular}
\endgroup
\caption{Comparison of simulated win rates ($\%$) and human annotated ``win + tie'' rates ($\%$) across models (Pearson $r = 0.975$, Kendall $\tau = 0.931$). The simulated win rates refer to the win rate predicted by the learned linear model on all 484 examples. The annotated ``win + tie'' rates refer to the human annotation on 100 examples from Figure \ref{fig: human_eval_win_rate}. }
\vspace{-3mm}
\label{table: simulated win rate}
\end{table}

\begin{table}[ht]
\centering
\captionsetup{font=footnotesize}
\footnotesize
\centering
\begin{tabular}{lc}
\toprule
\textbf{Model} & \textbf{Simulated($\%$)} \\ \midrule
Claude 3 Opus Direct & 77.5 \\
Claude 3 Opus Text-Augmented & 72.9 \\
Claude 3 Opus Self-Revision & 76.3 \\
LLaVA 1.6-7B Direct & 27.9 \\
LLaVA 1.6-7B Text-Augmented & 34.5 \\
LLaVA 1.6-7B Self-Revision & 32.9 \\
DeepSeek-VL-7B Direct & 25.8 \\
DeepSeek-VL-7B Text-Augmented & 37.4 \\
DeepSeek-VL-7B Self-Revision & 18.8 \\
Idefics2-8B Direct & 19.8 \\
Idefics2-8B Text-Augmented & 7.6 \\
Idefics2-8B Self-Revision & 3.3 \\
\bottomrule
\end{tabular}
\vspace{0.1in}
\caption{Simulated win rates for models that are not annotated by human, which are predicted by the learned linear model on all 484 examples.}
\label{table: simulated win rate for other models}
\end{table}

\begin{figure*}[t]
  \centering
  \includegraphics[trim={1cm 1cm 1cm 1cm},width=2\columnwidth]{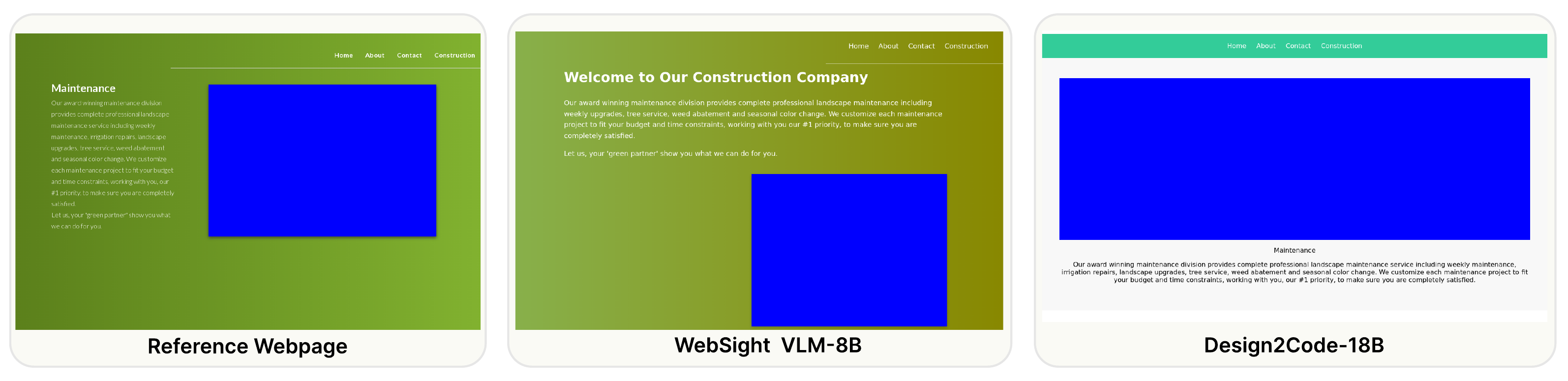}
  \caption{Comparison of WebSight VLM-8B and \texttt{Design2Code-18B}. WebSight VLM-8B excels at color recognition but hallucinates text contents.}
  \label{fig:open_source_models}
\end{figure*}

\section{Simulated Win Rate}
\label{appendix: simulate win rate}

Since we've shown that we can predict human judgment based on automatic metrics and achieve reasonable accuracy, we can also provide a simulated win rate using the learned coefficients and intercept. In practice, we remove the text similarity dimension and rerun the linear regression model on the same training examples (details provided in Appendix Table \ref{tab:coe analysis without text}). Using all 484 examples, we use the learned model to simulate the win rate and report the result in Table \ref{table: simulated win rate}. Since the learned model only predicts win/lose, we compare the simulated win rates with the annotated ``win + tie'' rates for fair comparison and observe a strong correlation between them, suggesting that our automatic metrics can also be aggregated into simulated win rates to facilitate model comparison. For models without annotated ``win + tie'' rates, we also provide the simulated win rate in Appendix Table \ref{table: simulated win rate for other models} for reference.

\section{What is the Learning Process of Different Dimensions?}
\label{sec:learning process}
We further plot the learning process of different automatic evaluation dimensions in Figure \ref{fig:skill_learning_ratio} to help us better understand the performance differences in Table \ref{tab:auto_eval}.
Specifically, we show the normalized performance of each aspect (so that $0$ before training and $1$ after training) for the base model checkpoint and all training checkpoints. On the one hand, performance on block-match, text, and position quickly saturate after training for $2000$ steps and remain stable afterward, possibly because these are the most transferable capabilities from the base model. On the other hand, the color similarity and the CLIP similarity steadily increase until $4000 - 5000$ steps. We assume that generating the correct color codes for texts and backgrounds benefits more from the HTML training data than other aspects and might be further improved by using the full Websight dataset and fully fine-tuning.

\section{More Case Study Examples}
\label{sec: qualitative Analysis}

By comparing WebSight VLM-8B vs \texttt{\modelname}-18B, we show a representative example in Figure~\ref{fig:open_source_models}, where WebSight VLM-8B is much better in coloring than \texttt{\modelname}-18B (color score $0.99$ vs $0.66$) and overall layout (position score $0.91$ vs $0.63$ and CLIP similarity $0.90$ vs $0.83$). However, WebSight VLM-8B tends to hallucinate texts and results in lower block-match ($0.85$ vs $0.99$) and text similarity scores ($0.98$ vs $1.0$). In general, we find that WebSight VLM-8B tends to have lower precision and recall than our model in terms of text matching.

\end{document}